\documentclass{isprs}
\usepackage{subfigure}
\usepackage{setspace}
\usepackage{geometry} 
\usepackage{enumerate}
\usepackage{natbib}
\usepackage{multirow}  
\usepackage{stfloats}
\usepackage{url}
\usepackage{amssymb}
\usepackage{amsmath}
\usepackage{float}
\usepackage{color}
\begin{document}

\title{Airborne LiDAR Point Cloud Classification with Graph Attention Convolution Neural Network}

\author{
Congcong Wen\textsuperscript{a,b,c}, Xiang Li\textsuperscript{c}, Xiaojing Yao\textsuperscript{a}, Ling Peng\textsuperscript{a}\thanks{Corresponding author. Email: plqiqi@126.com. }, Tianhe Chi\textsuperscript{a}
}

\address
{
\textsuperscript{a }Aerospace Information Research Institute, Chinese Academy of Sciences, Beijing, China.\\
\textsuperscript{b }
University of Chinese Academy of Sciences, Beijing, China.\\
\textsuperscript{c }Tandon School of Engineering, New York University, New York, United States. \\
}

\abstract
{
{\color{black}
Airborne light detection and ranging (LiDAR) plays an increasingly significant role in urban planning, topographic mapping, environmental monitoring, power line detection and other fields thanks to its capability to quickly acquire large-scale and high-precision ground information. To achieve point cloud classification, previous studies proposed point cloud deep learning models that can directly process raw point clouds based on PointNet-like architectures. And some recent works proposed graph convolution neural network based on the inherent topology of point clouds. However, the above point cloud deep learning models only pay attention to exploring local geometric structures, yet ignore global contextual relationships among all points. In this paper, we present a graph attention convolution neural network (GACNN) that can be directly applied to the classification of unstructured 3D point clouds obtained by airborne LiDAR. Specifically, we first introduce a graph attention convolution module that incorporates global contextual information and local structural features. The global attention module examines spatial relationships among all points, while the local attention module can dynamically learn convolution weights with regard to the spatial position of the local neighboring points and reweight the convolution weights by inspecting the density of each local region. Based on the proposed graph attention convolution module, we further design an end-to-end encoder-decoder network, named GACNN, to capture multiscale features of the point clouds and therefore enable more accurate airborne point cloud classification. Experiments on the ISPRS 3D labeling dataset show that the proposed model achieves a new state-of-the-art performance in terms of average F1 score (71.5\%) and a satisfying overall accuracy (83.2\%). Additionally, experiments further conducted on the 2019 Data Fusion Contest Dataset by comparing with other prevalent point cloud deep learning models demonstrate the favorable generalization capability of the proposed model.
}}


\maketitle

\section{Introduction}\label{Introduction}

Airborne light detection and ranging (LiDAR), as one of the most important techniques for data collection in earth observation (EO) systems, has the advantages of quickly acquiring large-scale and high-precision ground information, and plays an increasingly important role in urban planning \citep{yu2010automated}, topographic mapping \citep{krabill1984airborne, liu2008airborne, axelsson2000generation}, environmental monitoring \citep{huang2014wetland, bradbury2005modelling}, and power line detection \citep{sohn2012automatic, zhu2014fully}, etc. By employing airborne LiDAR for city scanning, a massive and irregular spatially distributed 3D point cloud with coordinates (X, Y, Z) and certain properties (e.g. intensity) can be acquired directly, the classification of which is an important research direction in the field of photogrammetry and remote sensing. However, achieving automatic  airborne LiDAR point cloud classification with high precision in real applications is challenging due to the high variability of object classes and complex object structure \citep{chen2007airborne, niemeyer2012conditional}.

Early research has mainly focused on solving the problem of airborne LiDAR point cloud classification by applying traditional machine learning-based models. These methods usually start with designing hand-crafted features, such as geometry features, radiometry features, topology features, echo features, and full waveform features, and then conduct point cloud classification by employing machine learning-based classifiers, including Support Vector Machine (SVM) \citep{zhang2013svm}, Adaboost \citep{lodha2007aerial}, Random Forest \citep{chehata2009airborne}, Markov Random Field \citep{munoz2009contextual, shapovalov2010nonassociative} and Conditional Random Field \citep{niemeyer2011conditional, niemeyer2014contextual}. Nevertheless, the calculation of these handcrafted features requires specific expert knowledge and has limited ability to extract effective features of the original point cloud data.

In recent years, deep learning models \citep{lecun2015deep} have drawn considerable attention from researchers due to their great success in various applications, such as natural language processing \citep{collobert2008unified},  speech recognition \citep{hinton2012deep}, time series prediction \citep{wen2019novel}, and image classification \citep{chan2015pcanet}, etc. Convolutional neural networks (CNN), one of the most prevalent models in deep learning can receive only regular inputs; therefore, early studies focusing on point cloud classification mostly transform point clouds to regular 3D voxels or collections of 2D feature images \citep{maturana2015voxnet, yang2017convolutional, yang2018segmentation, zhao2018classifying}. Nonetheless, this transformation process leads to inefficient computation and substantial memory consumption, and the transformation to 2D feature images causes spatial information loss.

 A deep learning model called PointNet, which can directly consume raw point clouds by exploiting multilayer perceptron (MLP) and max pooling to obtain the global feature representation \citep{qi2017pointnet}, is recently proposed for point cloud classifcation. The PointNet++ model, which first generates the partitioning of the point set and then builds a hierarchical neural network that employs PointNet recursively on a nested partitioning of the point cloud \citep{qi2017pointnet++}, is furhter presented. Similarly, various methods have been proposed to further improve the performance of point cloud classification by exploring local structure of point clouds \citep{li2018pointcnn, jiang2018pointsift, thomas2019kpconv}. More recently, a number of researchers introduce graph convolution neural network to classify point clouds based on the inherent topology of point clouds \citep{te2018rgcnn, wang2018local, wang2019dynamic}. Inspired by the visual attention mechanism, some work further present graph attention convolution network to learn adaptive local geometric structures and conduct point cloud classification \citep{wang2019graph, chen2019gapnet}.
 
However, the above point-based deep learning models in the field of computer vision only pay attention to explore local geometric structures, and ignore global contextual relationships among all points. In addition, few studies have investigated how to achieve airborne LiDAR point cloud classification by directly employing deep learning models to process raw point clouds \citep{yousefhussien2017fully, wang2018deep, wen2020directionally}. 


In this paper, we present a graph attention convolution neural network (GACNN) for airborne LiDAR point cloud classification. Specifically, the graph attention convolution module includes two types of attention mechanisms: a local attention module that combines edge attention and density attention, and a global attention module. The local edge attention module is designed to dynamically learn convolution weights in light of the spatial position relationships of neighboring points; thus, the receptive field of the convolution kernel can dynamically adjust to the structure of the point cloud. {\color{black}The local density attention module is devised to remedy the problem of the uneven density distribution of non-uniform sampled point cloud data. } Moreover, to learn global contextual information of the point clouds, we implement a global attention module by calculating Euclidean distance between every two individual points and utilize an MLP network to learn their attention weights. Furthermore, based on the presented graph attention convolution module, we develop an encoder-decoder network that can accept arbitrary sizes of input points and be trained in an end-to-end manner for airborne LiDAR point cloud classification. The key contributions of our work are summed up as follows:
\begin{enumerate}[1.]
\item 
We propose GACNN, a graph attention convolution neural network for airborne LiDAR point cloud classification that can be applied directly to raw point clouds to predict the semantic labels for arbitrarily sized input point clouds.

\item We design a local graph attention module that combines edge attention and density attention. The proposed edge attention module can adapt to the structure of the point cloud by dynamically adjusting kernel weights via learning from the local spatial layouts of neighboring points. The proposed density attention module can overcome problem of the uneven density distribution of non-uniform sampled point cloud data. 

\item We introduce a global graph attention module by taking the global spatial distribution among all points into consideration to capture global contextual features, which can further improve the performance of the proposed network.

\item The presented model realizes state-of-the-art classification performance on both ISPRS 3D Labeling Dataset and 2019 Data Fusion Contest Dataset.
\end{enumerate}

The remaining part of this paper is structured as follows. In Section 2,  a brief summary of point cloud classification methods is given with special regard to airborne LiDAR point clouds. We introduce the proposed GACNN in detail in Section 3. In Section 4, we conduct experiments to evaluate the performance of GACNN on the ISPRS 3D labeling benchmark dataset. The effect of the attention modules, and the superiority and generalization capability of the proposed model are discussed in Section 5. Finally, we conclude in Section 6.

\section{Related Work}\label{Related Work}

\subsection{Non-deep learning method}

Traditionally, studies accomplish airborne LiDAR point cloud classification by calculating handcrafted features and employing the classic machine learning models. \citeauthor{zhang2013svm} adopt the surface growing algorithm to cluster point clouds and utilize SVM to classify segments according to thirteen features related to radiometry, geometry, topology and echo characteristics \citep{zhang2013svm}. \citeauthor{lodha2007aerial} select five features, namely, image intensity, LiDAR return intensity, normal variation, height, and height variation, and implement AdaBoost to classify 3D airborne LiDAR data into four groups: buildings, trees, grass, and road \citep{lodha2007aerial}. \citeauthor{chehata2009airborne} apply Random Forests to select the most important features from multi-echo and full-waveform features, and classify airborne LiDAR point clouds on urban scenes \citep{chehata2009airborne}. However, these models ignore the contextual information of point clouds and predict the semantic label for each point individually, which creates noises in the classification results and inconsistency in the labels.

To resolve this issue, \citeauthor{niemeyer2011conditional} implement Conditional Random Field (CRF) method, which can be employed to incorporate contextual information and learn relations among objects, to classify airborne LiDAR point clouds based on geometrical and intensity features \citep{niemeyer2011conditional}. Besides, \citeauthor{niemeyer2014contextual} exploit ten types of features obtained from LiDAR point clouds and integrate Random Forest classifier into CRF to achieve reliable classification results, even in complicated urban scenes \citep{niemeyer2014contextual}.

Although the above non-deep learning methods achieve satisfying classification performance, they require manual calculation of features in advance, and the results are sensitive to the choice of features. 

\subsection{Deep learning method on grids or collections of images}
Deep learning, a novel machine learning method proposed in recent years, can automatically learn effective feature representations from a large amount of input data. CNNs, one of the most essential deep learning models, have made considerable progress in image classification tasks, such as object detection, semantic segmentation, and edge detection. However, because CNNs can handle only standard regular input format, direct application of CNNs to irregular and unordered 3D point clouds is infeasible.

To apply CNNs to airborne LiDAR point cloud classification, most researchers first convert the point cloud to regular 3D voxel grids or collections of images. \citeauthor{maturana2015voxnet} develop the Voxnet model to transform point clouds to 3D voxel grids, followed by a 3D CNN that predicts the semantic label based on the occupancy grid \citep{maturana2015voxnet}. However, 3D volumetric grids entail substantial memory consumption and computational cost. 

Therefore, some researchers have aimed to transform point clouds into collections of feature images, and subsequently exert CNNs to extract high-level representations of features and perform point cloud classification. \citeauthor{yang2017convolutional} generate 2D feature images by extracting the local geometric features, global geometric features and full-waveform features of each point;the features are then input into 2D CNNs for point cloud classification \citep{yang2017convolutional}. Moreover, \citeauthor{yang2018segmentation} change the previous image generation approach to a method that can implement the generation at different scales and design a multiscale CNN for final semantic classification based on five features, namely, intensity, eigenvalue features, planarity, sphericity, and variance of deviation angles \citep{yang2018segmentation}. \citeauthor{zhao2018classifying} propose a multiscale CNN that can automatically learn deep features of each LiDAR point by generating a set of contextual images from selected features of the LiDAR data, such as height, intensity and roughness \citep{zhao2018classifying}. However, these models cause spatial information loss and induce quantization error in the conversion process \citep{te2018rgcnn}.

\subsection{Deep learning method on point cloud}

As a pioneer work of directly applying deep learning models to raw point clouds, PointNet model \citep{qi2017pointnet} employ MLP to learn the features of individual points and a symmetric function (e.g. max pooling) to encode global information. Although PointNet provide a unified and efficient approach to 3D recognition tasks, it cannot capture the local structure of point cloud. To solve this problem, PointNet++ \citep{qi2017pointnet++} is developed by constructing a hierarchical neural network that applies PointNet recursively on partitioning of point cloud generated via sampling layers and grouping layers. 
Following these two models, many researchers proposed various deep learning models for point cloud classification based on PointNet-like architectures \citep{li2018pointcnn, jiang2018pointsift, thomas2019kpconv}.
 
Considering the inherent topological information of point clouds, researchers propose graph convolution neural network (GCNN) for point cloud classification applied on unordered 3D point clouds \citep{te2018rgcnn, wang2018local}. Similarly, \citeauthor{wang2019dynamic} design a dynamic graph convolution operations to capture local geometric structures by generating edge
features between a point and its neighborhoods \citep{wang2019dynamic}. More recently, some studies try to introduce the attention mechanism to learn a more adaptive local summary of the neighborhood. For instance, \citeauthor{chen2019gapnet} propose the GAPNet model, which employs a multi-head mechanism to aggregate attention features for each point from its neighborhoods and applies stacked MLP layers to capture local geometric features from raw point cloud \citep{chen2019gapnet}. \citeauthor{wang2019graph} introduce a graph attention convolution (GAC) to selectively focus on the most correlated part of the neighbors and train a graph attention convolution network for point cloud classification \citep{wang2019graph}. Note that these graph attention convolution neural networks just pay attention to local geometric structures, but our model also takes density distribution and global contextual relationships into consideration, which is the main difference of our graph attention convolution network from the previous ones in the computer vision field.

Specifically, in the field of airborne LiDAR point cloud classification, few studies apply deep learning models directly to raw point clouds. \citeauthor{yousefhussien2017fully} present a 1D-fully convolutional classification network that directly consumes 3D coordinates and three corresponding spectral features extracted from 2D georeferenced images for each point \citep{yousefhussien2017fully}. Furthermore, \citeauthor{wang2018deep} design a deep neural network with spatial pooling (DNNSP) by adopting a max pooling layer to aggregate the point-based features into the cluster-based features, which are then input to another MLP for point cloud classification \citep{wang2018deep}. In addition, our previous work explores a directionally constrained point convolution (D-Conv) module to extract local features of 3D point sets from the projected 2D receptive fields and then introduces a multiscale fully convolutional neural network based on the above module \citep{wen2020directionally}.

However, these point cloud deep learning methods focus on extracting local features and ignore the global relationships between individual points. In addition, the above models employ standard convolution kernels with a regular receptive fields, which neglect the structural connections between points and fails to account for varying point density.

In this paper, we propose a GACNN for airborne LiDAR point cloud classification that can directly handle unstructured 3D point clouds by considering local structural features and global contextual information simultaneously. Specifically, our proposed graph attention convolution module dynamically learns convolution weights and adaptively adjusts the convolution kernel according to the local structural connection of the point cloud. Moreover, the unbalanced density distribution of the point cloud is taken into account by our module.

\section{Methods}\label{Methods}
In this section, we first present the calculation of our proposed graph local attention (Section \ref{graph_local}) and global attention (Section \ref{graph_global}). Then, we describe how to combine these two attention mechanisms into our graph convolution module (Section \ref{graph_attention}). Based on the designed graph convolution module, we devise an encoder-decoder framework neural network (Section \ref{network}) that can enable learning multiscale features for airborne LiDAR point cloud classification.

\begin{figure}[ht]
\centering
\includegraphics[width=8cm]{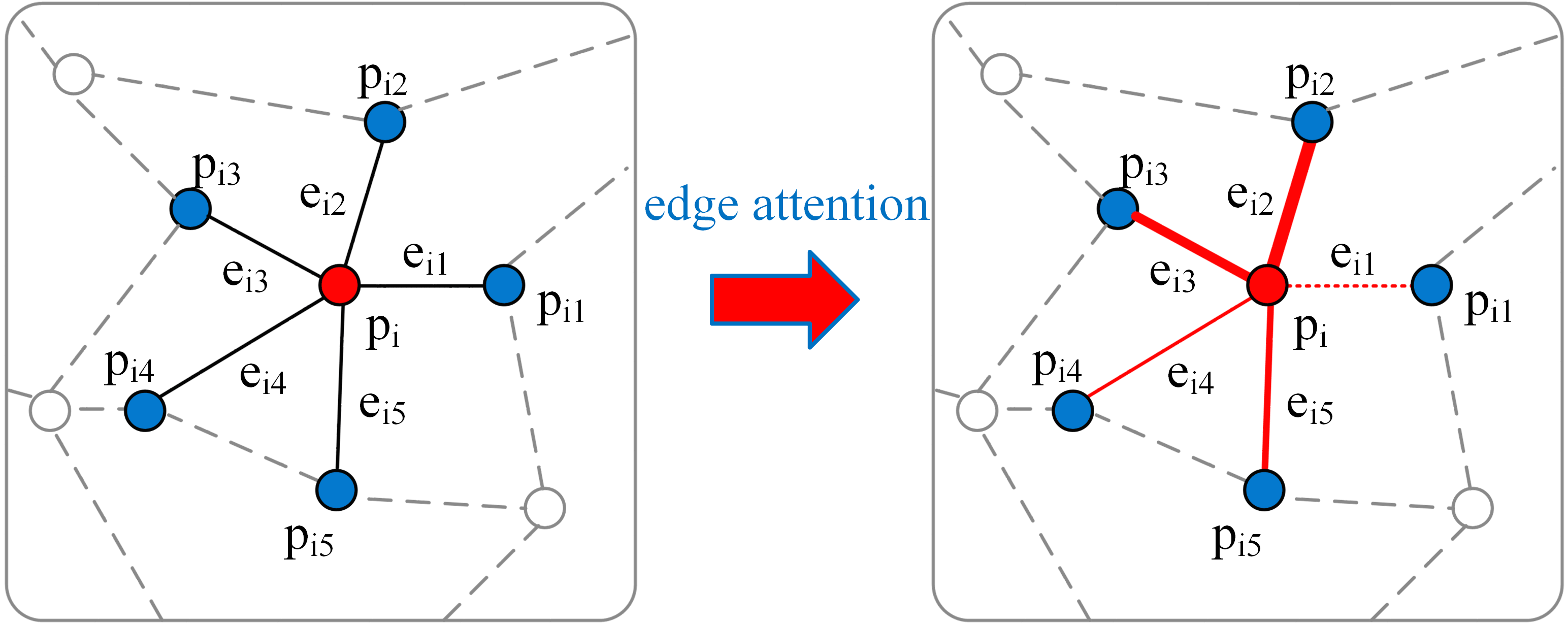}
\caption{  The effect of edge attention on standard graph convolution for a subgraph of point cloud. $x_{ij}$ indicates that $x_j$ is neighbor of point $x_i$ and $y_{ij}$ represents the corresponding edge between these two points. The learned convolution weights are strengthened or weakened after adding edge attention to the standard graph convolution (left). The thickness of the red line represents the magnitude of the attention coefficient, and the dotted red line indicates that the convolution weight is masked. }
\label{fig_edge_attention}
\end{figure}

\begin{figure*}[ht]
\centering
\includegraphics[width=16cm]{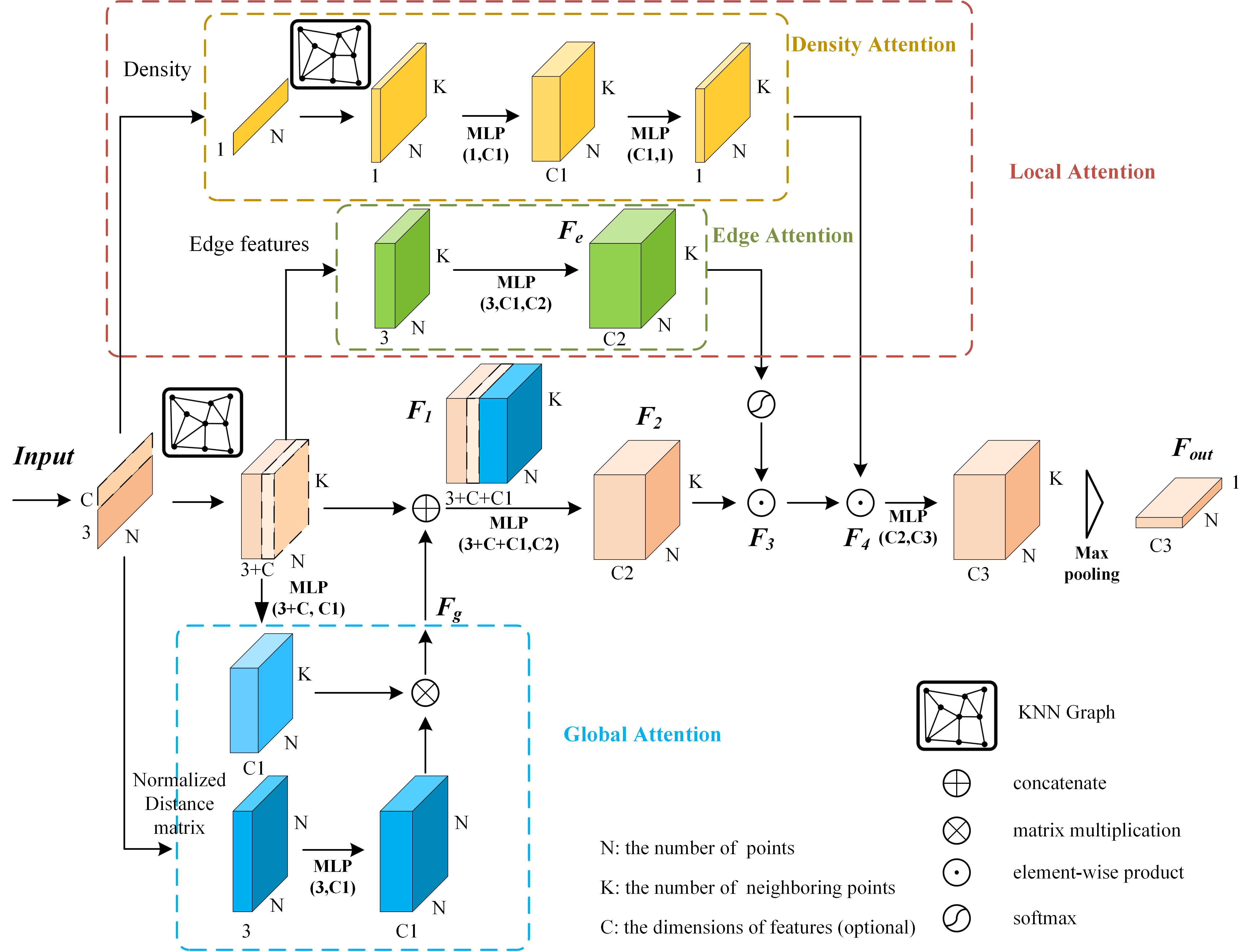}
\caption{
Illustration of our graph attention convolution module. Point clouds with 3D coordinates and optional features are input into our module. By calculating the k-nearest neighbor (KNN) graph according to the spatial position of each point, local neighboring point sets are generated, the features of which are concatenated with the global features calculated by the global attention module (blue dotted box). These concatenated features are fed into an MLP layer, the output of which is used to implement the element-wise product in conjunction with edge attention weights (green dotted box) and density attention weights (yellow dotted box), which are procured through the edge and density attention modules. By means of an MLP layer and max pooling, a feature map with the same amount of points as the input point cloud is finally obtained.
}
\label{fig_gac_module}
\end{figure*}

\subsection{Graph local attention}\label{graph_local}
To further explore the local structure of point clouds, two attention modules are added to standard graph convolution in this study, i.e. edge attention and density attention, so the proposed graph convolution module can dynamically learn convolutional kernel shapes to adjust to the structure of point sets and take the varying density distribution of non-uniform sampled point clouds into account.

\subsubsection{Edge attention }~\\
Given a point cloud $P = \{p_1, p_2, ... ,p_N\} \in \mathbb{R}^{N \times 3} $, a KNN graph $G(V, E)$ is  constructed according to its spatial neighbors, where $V = \{1, 2, ..., N\}$ are nodes of points, $E \subseteq |V| \times |\mathcal{N}_i| $ are the edges connecting pairs of points, and $\mathcal{N}_i$ is the neighborhood set of point $p_i$. Edge features are defined as $e_{ij} = (p_i - p_{ij})$, where $i \in V$, $j \in \mathcal{N}_i$, and $p_{ij}$ represents the neighboring point $p_j$ of point $p_i$.

To extract local structural features and learn the most related parts of the neighbors, the edge features are input into a nonlinear transformation function $\mathcal{F}e$, which is implemented via MLP and can be defined by :

\begin{equation}
\mathcal{F}e(e_{ij}) = \sum_{j=1}^{\mathcal{N}_i} {W_e^2}_{ij}* (\sum_{j=1}^{\mathcal{N}_i} {W_e^1}_{ij}*e_{ij} + b^1_i) + b^2_i
\label{eq_edge_mlp}
\end{equation}

where ${W_e}_{ij}$ and $b_i$, respectively, represent weight parameter and bias parameter of the MLP layer. 

In order to keep the same scale for the neighbor attention coefficients of different vertices, a softmax function was used to normalize weights across all neighbors to the reference vertex as follows:

\begin{equation}
\mathfrak{e_{ij}} = \frac{\exp(\mathcal{F}e(e_{ij}))}{\sum_{j=1}^{\mathcal{N}_i}\exp(\mathcal{F}e(e_{ij}))}
\label{eq_edge_norm}
\end{equation}

where $\mathfrak{e_{ij}}$ indicates the edge attention weight of vertex $p_j$ to reference vertex $p_i$. Figure \ref{fig_edge_attention} illustrates the effect of edge attention on standard graph convolution for a subgraph of a point cloud. After adding an edge attention mechanism, the proposed model can learn to strengthen or weaken convolution weight and dynamically adjust the actual receptive field of the convolution
kernel to the local structure of point cloud.

\subsubsection{Density attention}~\\
To overcome nonuniform density of the raw point cloud across different locations, we add density attention to the graph convolution module. First, the density of each point is estimated by the kernel density estimation (KDE), which can be represented as follows:

\begin{equation}
\hat{f}(p_i) =  \frac{1}{nh}\sum_{j=1}^{\mathcal{N}_i}K(\frac{p_i-p_{ij}}{h})
\label{eq_density}
\end{equation}

where K denotes the kernel function, n refers to the number of neighboring points, and h represents kernel window width. In this paper, we employ a Gaussian kernel function: 

\begin{equation}
\hat{f}(p_i) =  \frac{1}{nh(2\pi)^\frac{d}{2}}\sum_{j=1}^{\mathcal{N}_i}\exp(-\frac{1}{2}\left\|\frac{p_i-p_{ij}}{h}\right\|^2)
\label{eq_density}
\end{equation}

Then the inverse density of neighboring point $p_{ij}$ to reference point $p_i$ is computed and normalized by dividing the value by the maximum:

\begin{equation}
\mathcal{D}(p_{ij}) = \frac{1}{\hat{f}(p_{ij})}
\end{equation}

\begin{equation}
\mathcal{D}_{norm}(p_{ij}) = \frac{\mathcal{D}(p_{ij})}{\max \limits_{j=1,2,...,\mathcal{N}_i}\mathcal{D}(p_{ij})}
\label{eq_density}
\end{equation}

Finally, the density attention weight of point $x_{ij}$ to reference point $x_i$ is calculated by MLP, as shown in Eq.\ref{eq_density_attention}.

\begin{equation}
\mathfrak{d_{ij}} = \sum_{j=1}^{\mathcal{N}_i} {W_d^2}_{ij}*(\sum_{j=1}^{\mathcal{N}_i} {W_d^1}_{ij}*\mathcal{D}_{norm}(p_{ij}) + b^1_i) + b^2_i
\label{eq_density_attention}
\end{equation}

\subsection{Graph global attention}\label{graph_global}
Previous works focus on extracting local features and obtaining global information by employing a symmetric function but ignore the spatial relationships among all points. To remedy this problem, we propose a graph global attention module to learn global contextual information of the point cloud. The Euclidean distance between every two individual points in each coordinate direction is calculated to obtain the distance matrix with dimension $N*N*3$, which can be represented as:

\begin{equation}
\mathbb{D} = \sum_{i=1}^{N} \sum_{j=1}^{N} \left[ p^i_x-p^j_x,  p^i_y-p^j_y,   p^i_z-p^j_z\right]^T
\label{eq_global_distance}
\end{equation}

Similarly, we use a softmax layer to normalize the distance for each point, which can be expressed as:

\begin{equation}
d_{ij} = \frac{\exp(\mathbb{D}_{ij})}{\sum_{j=1}^{N}\exp(\mathbb{D}_{ij})}
\label{eq_dis_norm}
\end{equation}

where $\mathbb{D}_{ij}$ and $d_{ij}$ signify the distance and normalized distance between points $p_i$ and $p_j$, respectively. Another MLP layer is employed as nonlinear transform function to obtain the final attention weight from the normalized distance, which is given by:

\begin{equation}
\mathfrak{g_{ij}} = \sum_{j=1}^{N} {W_g}_{ij}*d_{ij} + b_i
\label{eq_global_attention}
\end{equation}

\subsection{Graph attention convolution neural network}\label{GACNN}

In the last two sections, we introduce the local and global attention modules in detail. In this section, we first present how the two attention mechanisms are fused into the proposed graph attention convolution module. Then, we develop an end-to-end encoder-decoder network to learn multiscale features from raw point clouds.

\begin{figure*}[ht] 
\centering
\includegraphics[width=17cm]{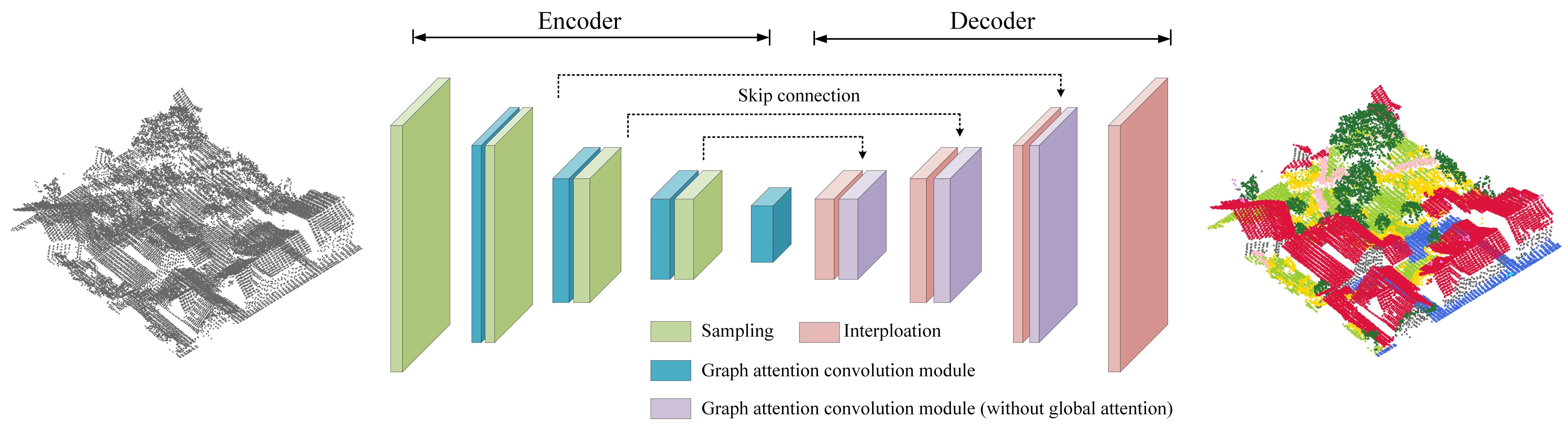}
\caption{
Illustration of the graph attention convolution neural network (GACNN) architecture. In the encoder network, the sampling layer and the proposed graph attention convolution module are recursively employed to extract multiscale features. To propagate the learned features to original points, the subsampled features are interpolated and concatenated with the corresponding encoder features, followed by the implementation of our graph attention convolution module without global attention in each decoder. After implementing the same number of decoder layers as encoder layers, the final semantic label for each point is predicted by 1 $\times$ 1 convolution.
}
\label{fig_gacnn}
\end{figure*}

\subsubsection{Graph attention convolution module}\label{graph_attention}~\\
As illustrated in Figure \ref{fig_gac_module}, a point cloud $P = \{p_1, p_2, ... ,p_N\} \in \mathbb{R}^{N \times 3} $ and its corresponding features  $F = \{f_1, f_2, ... ,f_N\} \in \mathbb{R}^{N \times C} $  are input into module. Note that these features are optional (represented by black dashed lines in Figure \ref{fig_gac_module}), and only the 3D coordinates are taken as input (i.e. $C = 0$) when these additional features are not required. The KNN graph is built according to coordinates of the point cloud, and local neighboring point sets $\mathcal{G} = \{g_i \in \mathbb{R}^{K \times (3+C)}, i = 1, 2,...,N \} \in \mathbb{R}^{N \times K \times (3+C)}$ are obtained. 

As discussed in Section \ref{graph_global}, we first calculate the normalized Euclidean distance, and then learn a nonlinear transformation $\mathbb{R}^{N \times N \times 3} \to \mathbb{R}^{N \times N \times C1}$, which can be implemented via MLP to procure global attention weights. Meanwhile, another MLP is employed to learn a mapping: $\mathbb{R}^{N \times K \times (3+C)} \to \mathbb{R}^{N \times K \times C1}$ from the local neighboring point sets of the input point cloud. The global attention feature map $ \boldsymbol{F}_g \in \mathbb{R}^{N \times K \times C1}$ is generated via matrix multiplication between global attention weights and the mapping result of the local neighboring point sets. Finally, after concatenation of the local neighboring point sets and $\boldsymbol{F}_g$, the feature map $ \boldsymbol{F}_{1} \in \mathbb{R}^{N \times K \times (3+C+C1) }$ is obtained. The above process describes how global attention is added  to our graph attention convolution module, which can be found in the blue dotted box in Figure \ref{fig_gac_module}.

Next, we introduce how local attention is added to the graph attention convolution module, as shown in the red dashed box in Figure \ref{fig_gac_module}. Based on the spatial coordinates of the local point sets, we calculate edge features $E \in \mathbb{R}^{N \times K \times 3} $, followed by two shared MLP layers (C1,C2) to output to the local edge attention feature map $\boldsymbol{F}_e \in \mathbb{R}^{N \times K \times C2}$ (see green dotted box in Figure \ref{fig_gac_module}). Then, the edge attention weights are acquired by applying the softmax layer to normalize $\boldsymbol{F}_e$. Thereafter, element-wise product of the edge attention weights and the feature map $ \boldsymbol{F}_{2} \in \mathbb{R}^{N \times K \times C2}$, which is the output of an MLP applied to $\boldsymbol{F}_{1}$, is performed to generate feature map $ \boldsymbol{F}_{3} \in \mathbb{R}^{N \times K \times C2}$. 
Regarding density attention (see the yellow dotted box in Figure 3), the density of each input raw point is calculated, and the density of the local neighboring point sets, which are selected through the KNN graph pursuant to spatial position, are aggregated to constitute a density neighboring matrix. Then, two shared MLP layers (C1,1) are used to acquire the density attention weights via element-wise product with $\boldsymbol{F}_{3}$. The output of the element-wise product is feature map  $ \boldsymbol{F}_{4} \in \mathbb{R}^{N \times K \times C2}$, which is again input into another MLP layer and max pooling layer to secure the feature map $ \boldsymbol{F}_{out} \in \mathbb{R}^{N \times C3}$, which is final output of our graph attention convolution module.

\begin{figure*}[ht] 
\centering
\includegraphics[width=16cm]{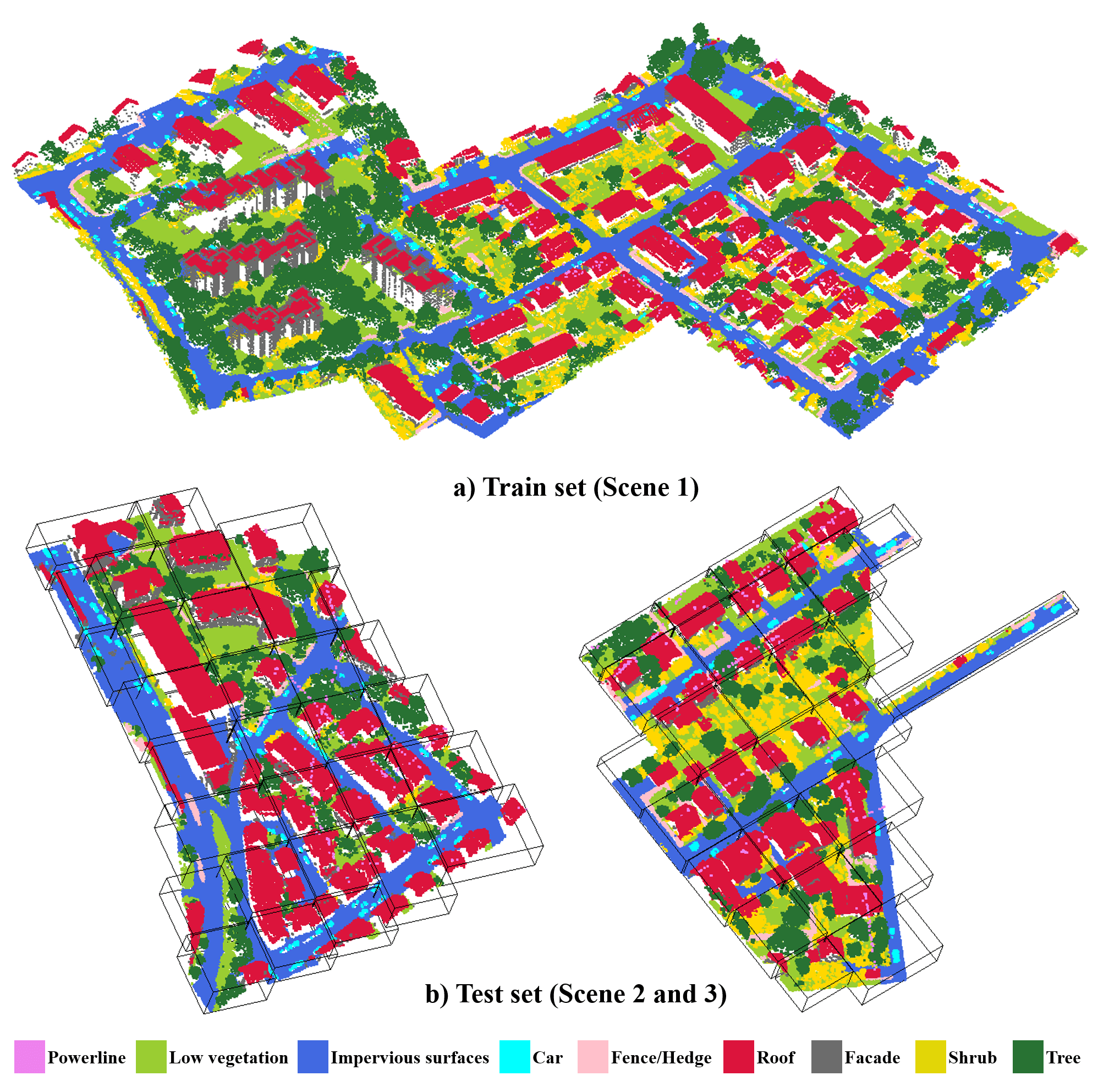}
\caption{
Three scenes of the ISPRS 3D labeling dataset. Scene 1 (top) is employed as the training set, and Scenes 2 and 3 (bottom) are used as the test set. The legend at the bottom defines the color representing each category. Best viewed in color.
}
\label{fig_dataset}
\end{figure*}

\subsubsection{Overall architecture}\label{network}~\\
Inspired by SegNet \citep{badrinarayanan2017segnet} and PointNet++ \citep{qi2017pointnet++}, we develop an encoder-decoder GACNN in an end-to-end manner based on the above graph attention convolution module for airborne LiDAR point cloud classification. Raw point clouds with 3D coordinates and optional features are directly input into our encoder network. Subsequently, the sampling layer implemented via the farthest point sampling algorithm, and graph attention convolution module are employed recursively four times to extract multiscale features in the encoder network. To propagate the learned features from the encoded sampled points to the original points, the interpolation is first attained through inverse distance weighting within the decoder network. More details of the sampling and interpolation can be found in the PointNet++ \citep{qi2017pointnet++}. Then the interpolated features are concatenated in a skip manner with the point features from the corresponding encoder stages. Next the concatenated features are input into our graph attention convolution module to capture features from the coarse-level information. Note that the graph attention convolution module contains only two local attention mechanisms in each decoder. After the last interpolation, the feature collections encompass the same number of points as that of the original point sets and are fed into a 1 $\times$ 1 convolution to obtain the final semantic label for each point.

\section{Experiments}\label{Experiments}
In this section, the performance of our GACNN for airborne LiDAR point cloud classification is evaluated on a real-world dataset. We briefly describe the experimental dataset in Section \ref{Dataset} and introduce the evaluation metrics of the point cloud classification in Section \ref{Evaluation metric}. The implementation details of our GACNN model are  illustrated in Section \ref{Implementation Details}. In Section \ref{Classification results}, we present the classification results of the GACNN model on the dataset.  

\subsection{Dataset}\label{Dataset}

We assess the performance of our model on the International Society for Photogrammetry and Remote Sensing (ISPRS) 3D labeling dataset, which is composed of airborne laser scanning data acquired with a Leica ALS50 system from Vaihingen, Germany \citep{cramer2010dgpf}. The LiDAR data are categorized into nine semantic classes \citep{niemeyer2014contextual}, that is power line, low vegetation (low\_veg), impervious surface (imp\_surf),  fence/hedge, car, roof, facade, shrub, and tree. Scenes from three different areas are provided on the ISPRS 3D labeling website: one scene with 753,876 points is employed as the training set and the other two scenes with 411,722 points are used as the test set. Table \ref{table_Dataset_number} describes the number of points of each category for each set. The spatial XYZ coordinates, intensity, the return of number, the number of returns and semantic labels are provided for each set. Figure \ref{fig_dataset} shows the three scenes in the dataset, and the rendering color of each category is referenced to \citep{blomley20163d}.

\begin{table}[h]
\begin{center}
\caption{The number of points in each category for the training set and test set}
\label{table_Dataset_number}
\begin{tabular}{l c c}

\hline
Categories & Training Set & Test Set
\\
\hline
Powerline & 546  & 600\\
Low vegetation & 180,850 & 98,690\\
Impervious surfaces & 193,723 & 101,986\\
Car &  4,614 & 3,708\\
Fence/Hedge & 12,070 & 7,422\\
Roof & 152,045 & 109,048\\
Facade & 27,250 & 11,224\\
Shrub & 47,605 & 24,818\\
Tree & 135,173 & 54,226\\
\hline
Total & 753,876 & 411,722\\       
\hline
\end{tabular}
\end{center}
\end{table}

\begin{figure*}[ht] 
\centering
\includegraphics[width=15cm]{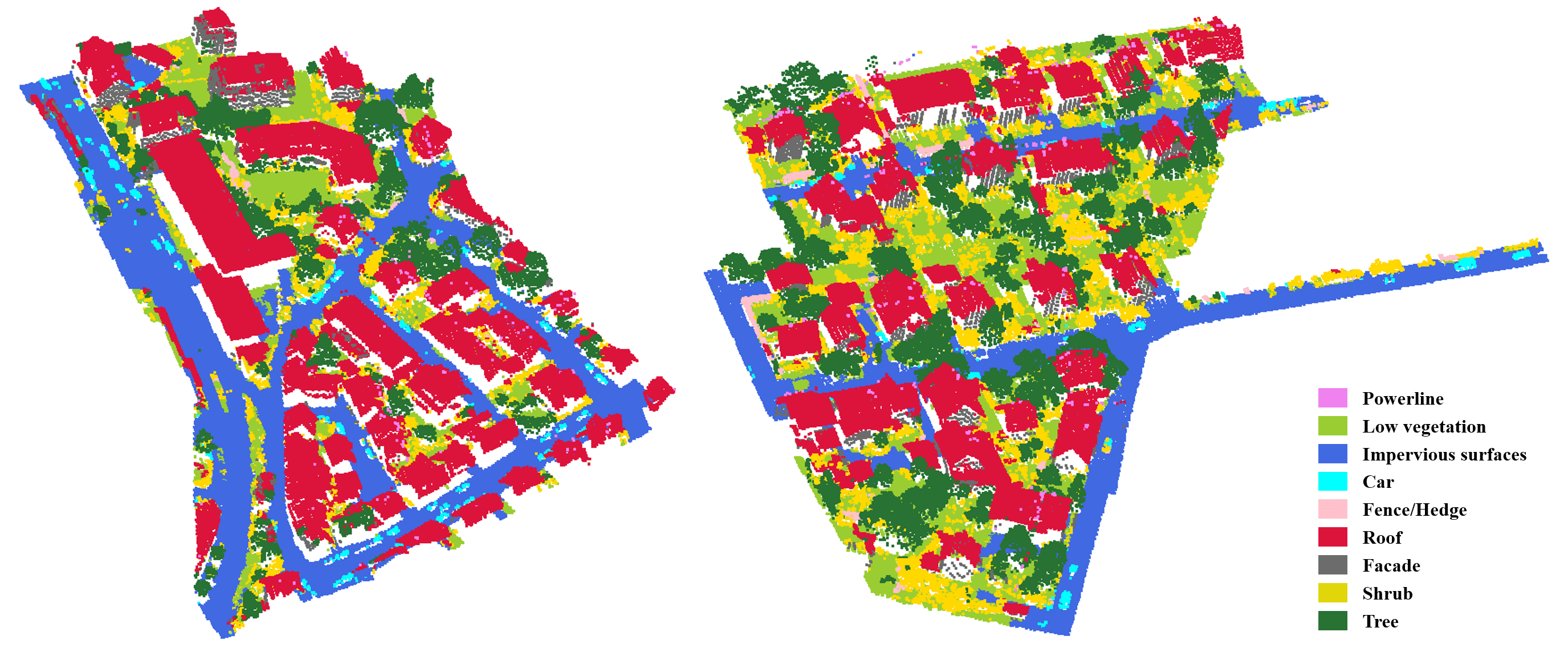}
\caption{
The classification results of our GACNN model on test set (Scene 2 and 3) of ISPRS dataset. The legend in the lower right corner represents the color corresponding to each category. Best viewed in color.
}
\label{fig_result}
\end{figure*}

\begin{figure*}[h] 
\centering
\includegraphics[width=15cm]{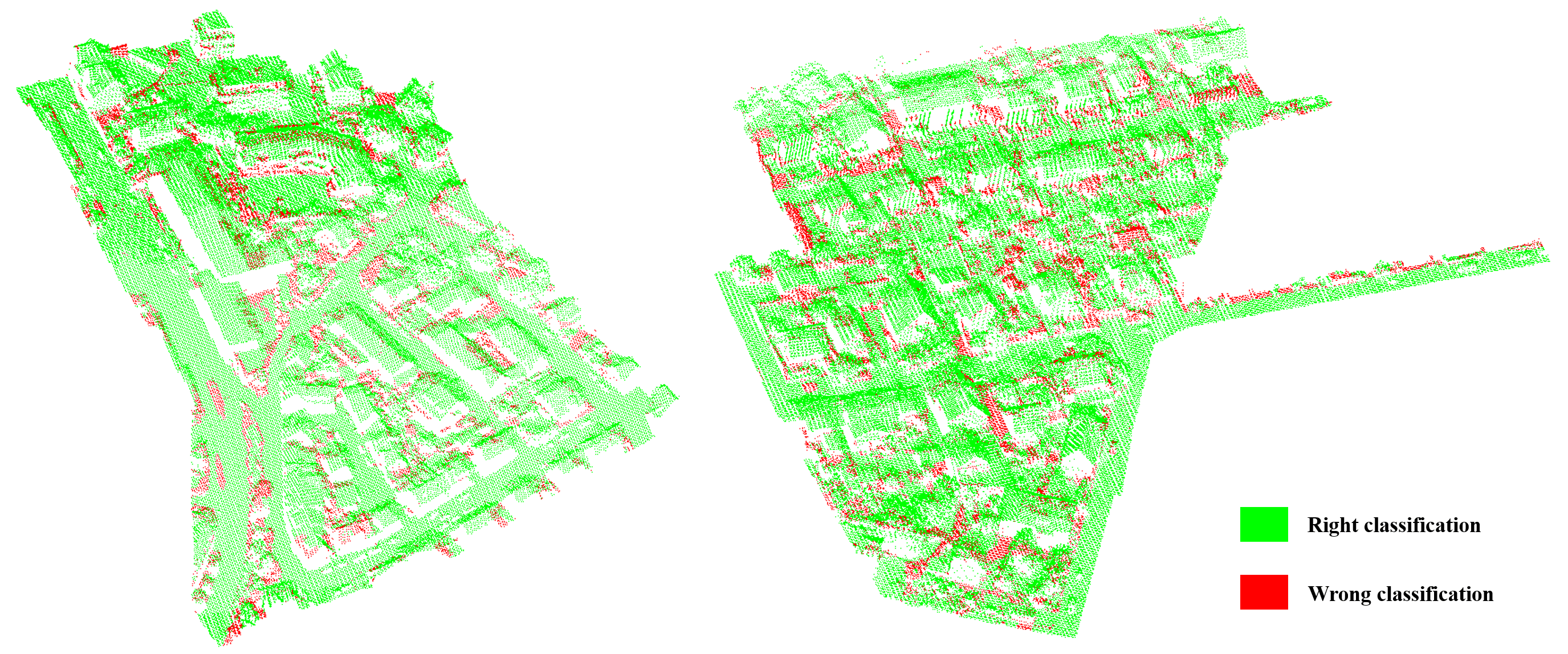}
\caption{
The classification error map of our GACNN model on test set (Scene 2 and 3) of ISPRS dataset. The points marked in green and red represent the correct and incorrect classification results, respectively. Best viewed in color.
}
\label{fig_errormap}
\end{figure*}

\begin{table*}[h]
\begin{center}
\caption{The classification confusion matrix of the proposed graph attention convolution neural network (GACNN) model. Precision/correctness, recall/completeness, and F1 score are also reported. The overall accuracy and average F1 score for the classification results are 83.2 \% and 71.5 \%, respectively. }
\label{table_confusion_matrix}
\begin{tabular}{l c c c c c c c c c}

\hline
Categories & power & low\_veg & imp\_surf & car & fence\_hedge & roof & facade & shrub & tree 
\\
\hline
power & \textbf{0.775} & 0.000 & 0.000 & 0.000 & 0.000 & 0.152 & 0.002 & 0.005 & 0.067 \\
low\_veg & 0.000 & \textbf{0.780} & 0.082 & 0.001 & 0.006 & 0.008 & 0.004 & 0.102 & 0.017 \\
imp\_surf & 0.000 & 0.054 & \textbf{0.942} & 0.000 & 0.000 & 0.000 & 0.001 & 0.003 & 0.000 \\
car & 0.000 & 0.040 & 0.012 & \textbf{0.709} & 0.039 & 0.020 & 0.010 & 0.152 & 0.019 \\
fence\_hedge & 0.000 & 0.091 & 0.015 & 0.009 & \textbf{0.290} & 0.026 & 0.026 & 0.440 & 0.103 \\
roof & 0.001 & 0.019 & 0.000 & 0.000 & 0.001 & \textbf{0.912} & 0.013 & 0.013 & 0.041 \\ 
fac & 0.002 & 0.056 & 0.005 & 0.003 & 0.005 & 0.151 & \textbf{0.540} & 0.134 & 0.104 \\
shrub & 0.000 & 0.107 & 0.006 & 0.005 & 0.027 & 0.045 & 0.025 & \textbf{0.611} & 0.174 \\ 
tree & 0.000 & 0.015 & 0.000 & 0.001 & 0.004 & 0.022 & 0.011 & 0.145 & \textbf{0.802} \\
\hline
Precision/Correctness & 0.746 & 0.860 & 0.919 & 0.860 & 0.544 & 0.951 & 0.648 & 0.378 & 0.776 \\
Recall/Completeness & 0.775 & 0.780 & 0.942 & 0.709 & 0.290 & 0.912 & 0.540 & 0.611 & 0.802 \\
F1 score & 0.760 & 0.818 & 0.930 & 0.777 & 0.378 & 0.931 & 0.589 & 0.467 & 0.789 \\

\hline

\end{tabular}
\end{center}
\end{table*}

\subsection{Evaluation metric}\label{Evaluation metric}

According to the standard evaluation metrics of the ISPRS 3D labeling contest, precision, recall, and F1 score, overall accuracy (OA) are used to evaluate the performance of point cloud classification. In general, OA, which is specified as the percentages of correctly classified points in the test set, examines the classification accuracy for all categories. F1 score measures the classification performance for each category based on the precision and recall of the classification model and is a better choice than OA when there is a large difference in the number of points in each category. The OA, precision, recall and F1 score are designated as follows:

\begin{equation}
OA = \frac{TP+TN}{TP+TN+FP+FN}
\end{equation}

\begin{equation}
precision = \frac{TP}{TP+FP}
\end{equation}

\begin{equation}
recall = \frac{TP}{TP+FN}
\end{equation}

\begin{equation}
F1 = 2*\frac{precision*recall}{precision+recall}
\end{equation}

where TP, TN, FP and FN, respectively, denotes true positive (the fraction of positives correctly classified), true negative (the fraction of negatives correctly classified), false positive (the fraction of positives misclassified), and false negative (the fraction of negatives misclassified). 

\subsection{Implementation Details}\label{Implementation Details}

As stated in section \ref{GACNN}, the raw 3D coordinates and optional features (intensity values acquired from the airborne LiDAR data and the height above ground features are adopted in this paper) of the point cloud are input into our model; however, the entire training point set cannot be directly fed into the network because of the limited GPU memory. To resolve this problem, we separated the whole scene into a set of 30 m*30 m*40 m cuboid regions. During the model training step, to make the model more robust and prevent overfitting, we first randomly select several cuboid regions, randomly choose 8,192 points from each cuboid region and indiscriminately drop 12.5\% of these points. In terms of the model testing stage, all points from split cuboid regions in the test set (see Figure. \ref{fig_dataset} b)) are fed into the trained model to achieve point-to-point classification, although the number of points is different for each cuboid region. Note that the split small broken regions that occur at the edges of the scene are merged into the surrounding larger regions to ensure the integrity of each cuboid region.

The proposed model is implemented based on the Tensorflow framework. We employ the ADAM optimizer with a learning rate of 0.01 and divide the learning rate by 2 every 3,000 steps. The batch size is set to 8 for model training and to 1 for model testing. In the encoder network, the number of sampling points in each sampling layer is set to 1024, 512, 64, and 16, with the feature dimensions C1, C2, and C3 (see Figure \ref{fig_gac_module}) in each proposed graph attention convolution module set to (32, 32, 64), (64, 64, 128), (128, 128, 256) and (256, 256, 512), and the number of neighboring points K set to 32. In the decoder network, the feature dimensions C2 and C3 (without global attention in the decoder) in each proposed graph attention convolution module are (512, 512), (256, 256), (256, 128) and (128, 128), individually, and the number of neighboring points K is 16. Note that the model parameters are determined through a series of comparative experiments; we do not elaborate on the experimental details since that is not the focus of our paper.

\begin{figure}[h]
\centering
\includegraphics[width=8cm]{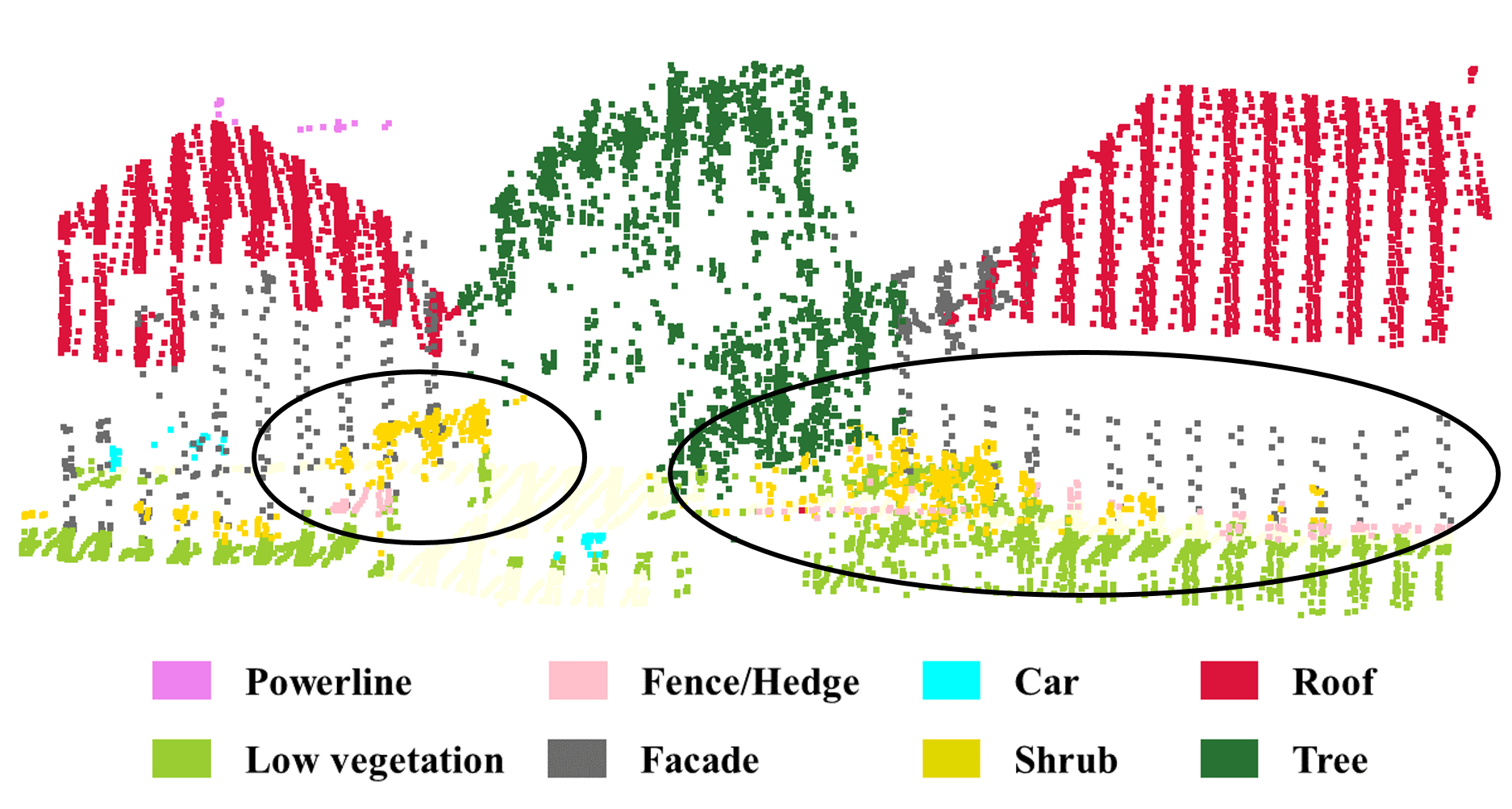}
\caption{Illustration of the mingling of the shrub category points with other category points. The shrub category points (colored with gold) are mixed with fence/hedge category points (colored with pink) marked by the left and right black circle, and tree category points (colored with emerald-green) marked by the right black circle. }
\label{fig_data_shrub}
\end{figure}

\subsection{Classification results}\label{Classification results}

As covered in the last section, the model training is conducted on the ISPRS dataset for 1000 epochs, taking 10 hours on a Titan Xp GPU until convergence. Subsequently the point cloud of each cuboid region of the test set (Scene 2 and 3) is directly input into the trained model in turn to predict the semantic labels of each point. Figure \ref{fig_result} and Figure \ref{fig_errormap} show the classification results and error maps of our GACNN model which correctly labelled the majority of the points in the test set. 

To quantitatively assess the classification performance of our model for the points in each category, the classification confusion matrix, precision, recall and F1 score of each category are all calculated and listed in Table \ref{table_confusion_matrix}. Table \ref{table_confusion_matrix} shows that the proposed model achieves satisfactory classification (F1 score greater than 75\%)  on six categories, namely, power line, low vegetation, impervious surfaces, car, roof, and tree. Acceptable classification performance is achieved on facade categories due to the mingling of facade category with the roof, shrub and tree categories, which causes the misclassification of facade points.

However, our model demonstrates poor classification performance for the fence/hedge and shrub categories, most likely because of the similar geographical distribution and topological features of these two categories. In addition, Table \ref{table_confusion_matrix} shows that some points of the shrub category were misclassified as tree, which may also result from the mixing and lack of obvious boundaries between the shrub and tree categories. As shown in Figure \ref{fig_data_shrub},  points of the shrub category are mixed with points of the fence/hedge category (marked with the left circle and right circle) and points of the tree category (marked with the right circle).

\section{Discussion}\label{Discussion}
In this section, we examine the effect of the proposed global and local attention modules through a set of ablation studies in section \ref{ab_attention}. In Section \ref{comparisions}, we compare the proposed GACNN with other state-of-the-art methods for airborne LiDAR point cloud classification on the ISPRS dataset. Moreover, we discuss the generalization capability of GACNN on 2019 Data Fusion Contest Dataset.

\subsection{Ablation study for attention modules}\label{ab_attention} 

\begin{figure*}[htb] 
\centering
\includegraphics[width=16cm]{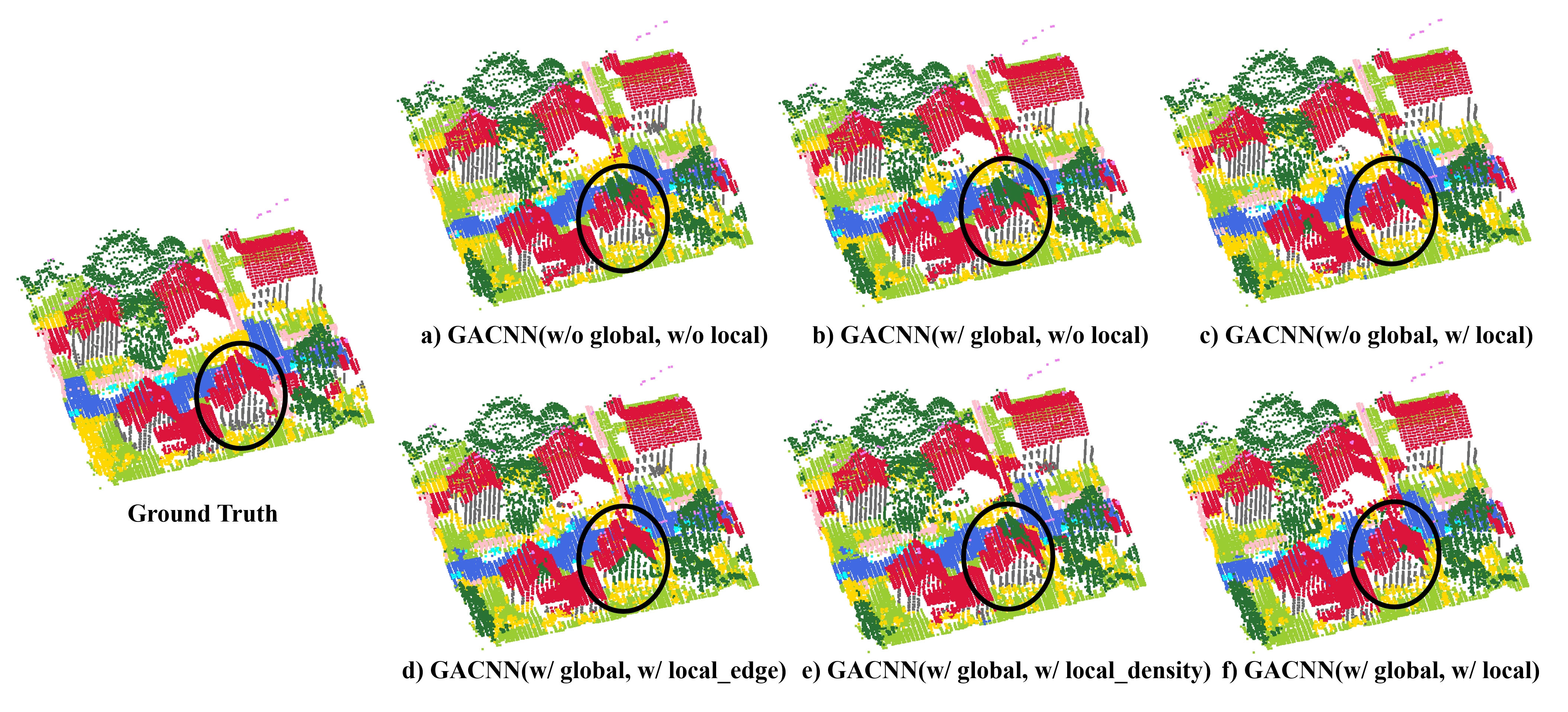}
\caption{
The classification results of our proposed GACNN model with different attention modules on a randomly selected region. The black circle highlights the most obvious differences in the classification results obtained by the six models.
}
\label{fig_effect_attention}
\end{figure*}

\begin{figure*}[htb] 
\centering
\includegraphics[width=16cm]{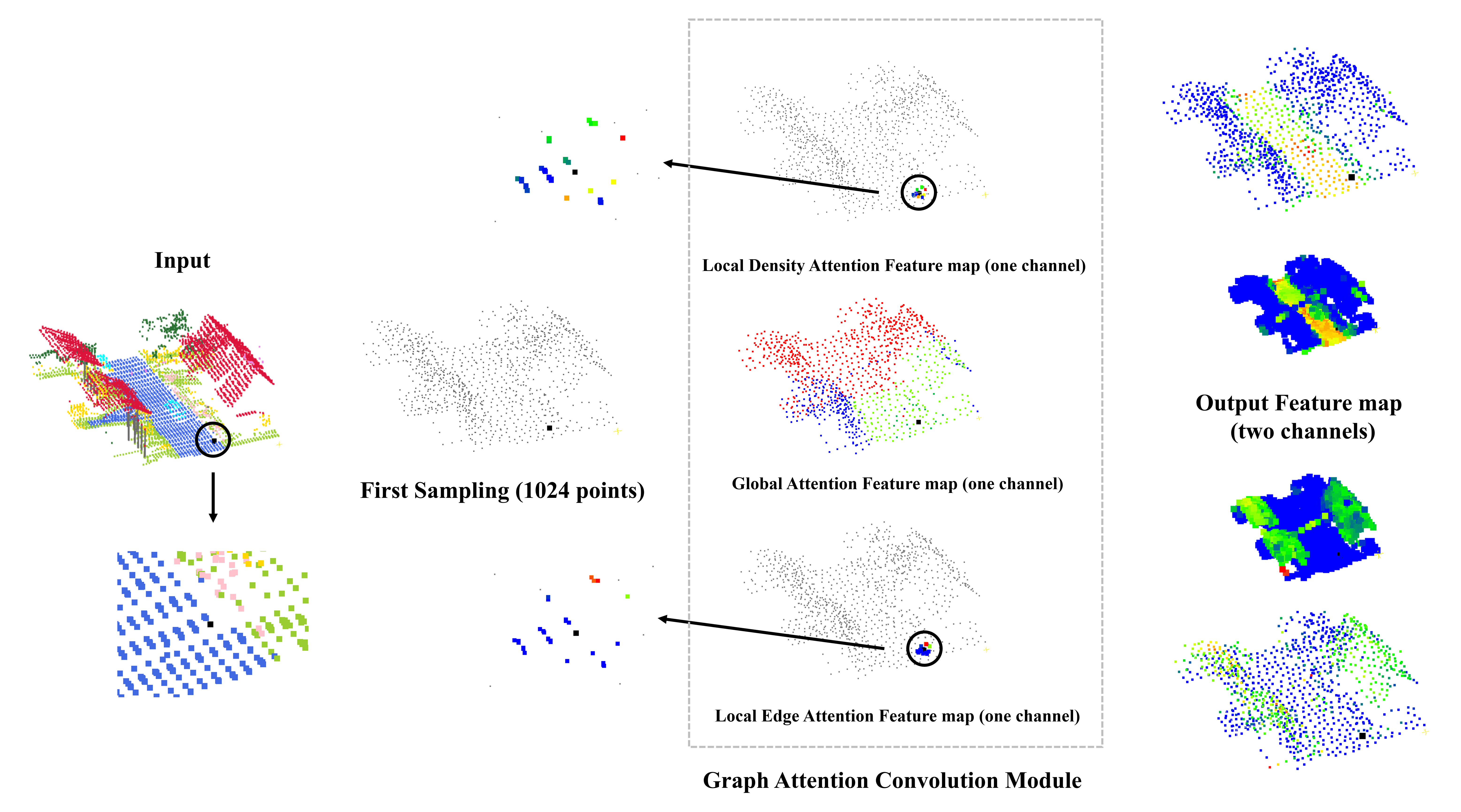}
\caption{
Visualized feature maps of the global and local attention modules in the first encoder layer of GACNN model on a randomly selected point in a region. The black circles represent the partial enlargement of the neighbors of the selected point. The colors of the point cloud represent different features (except for the input point cloud).
}
\label{fig_show_attention}
\end{figure*}

To evaluate the effectiveness of the proposed graph attention convolution module (i.e. the global attention module and local attention module), we design an ablation experiment to compare six models: a) GACNN model without the global and the local attention modules (w/o global, w/o local), b) GACNN model with the global attention module but without the local attention modules (w global, w/o local), c) GACNN model without the global attention module but with the local attention modules (w/o global, w/ local), d) GACNN model with the global and the local edge attention module (w/ global, w/ local\_edge), e) GACNN model with global and local density attention module (w/ global, w/ local\_density), and f) GACNN model with the global and the two local attention modules (w/ global, w/ local).

Table \ref{table_effect_attention} shows the classification results of the above six models. 
{\color{black}
Take local attention module as an example, it can be found that adding local attention module helps to improve the classification performance by comparing a) GACNN with c) GACNN model, and b) GACNN with f) GACNN model. To be more specific, the comparison of classification results between b) GACNN and d) GACNN model, as well as e) GACNN and f) GACNN model demonstrate the effectiveness of the first local module, namely edge attention module. Similarly, the value of the second local module, density attention module, can be illustrated by comparing b) GACNN with e) GACNN model, and d) GACNN with f) GACNN. We can implement alike analysis to verify the effectiveness of global attention module.  Trough these comparisons, it can be found that each attention module helps to improve the classification performance to some extent.} 

To more intuitively demonstrate the effects of each attention module, we randomly select a region and plot the classification results of the six models, as shown in Figure \ref{fig_effect_attention}. The three models without the local edge attention module (a, b and e) misclassify the roof category points as tree category points, which validates our assumption that the addition of the local edge attention module promotes the classification performance by dynamically learning the convolution kernel according to the local structure of the point cloud.

Furthermore, Figure \ref{fig_show_attention} visualizes the feature maps of the global and local attention modules in the first encoder layer of the GACNN model on a randomly selected region. The input point cloud on this region is first dowsampled into 1024 points and then fed into our graph attention convolution module. Three attention feature maps of the sampled points are procured by implementing the procedures illustrated in Figure \ref{fig_gac_module}. The black circles and arrows in Figure \ref{fig_show_attention} show the partial enlargement of the local neighboring points of the selected point. The learned local features of the two local attention modules correspond to the label of the neighboring points in general. Moreover, the global attention feature map shows that the features of the neighboring points differ when the distance between the selected point and the surrounding points varies. Finally, the output feature maps (two channels) of the graph attention convolution module are shown on the right-most side of Figure \ref{fig_show_attention}. The features of impervious surfaces and roofs are substantially captured by our graph attention convolution module, which further authenticates the effect of the global and local attention modules.

\begin{table}[h]
\begin{center}
\caption{The classification overall accuracy (OA) and average F1 score of our proposed GACNN model with different attention modules on the ISPRS benchmark dataset. The boldface text indicates the model with the highest performance.}
\label{table_effect_attention}
\begin{tabular}{l c c}
\hline
Model & OA & Average F1
\\
\hline
a) GACNN(w/o global, w/o local) & 0.816 & 0.697 \\
b) GACNN(w/ global, w/o local)  & 0.828 & 0.707 \\
c) GACNN(w/o global, w/ local) & 0.824 & 0.705 \\
d) GACNN(w/ global, w/ local\_edge) & 0.831 & 0.709 \\
e) GACNN(w/ global, w/ local\_density) & 0.830 & 0.710 \\
\textbf{f) GACNN(w/ global, w/ local)} & \textbf{0.832} & \textbf{0.715} \\
\hline
\end{tabular}
\end{center}
\end{table}

\subsection{Comparisons with other methods}\label{comparisions}

\begin{figure*}[h] 
\centering
\includegraphics[width=16cm, height=9cm]{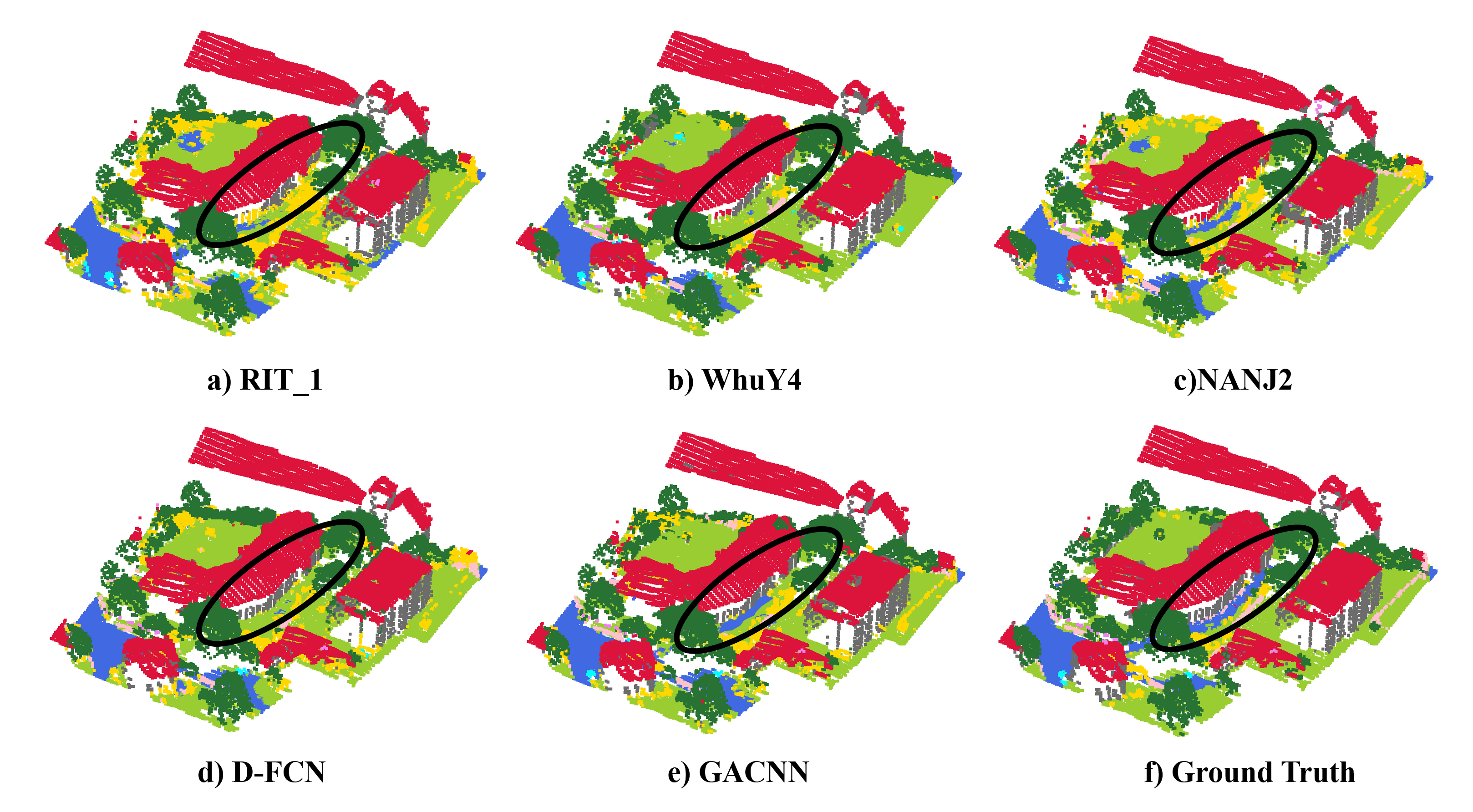}
\caption{
The classification results of RIT\_1 model, WhuY4 model, NANJ2 model, D-FCN model, GACNN model and ground truth on a selected complicated scene area. The black circle indicates the most obvious differences in the classification results obtained by the six models. Best viewed in color.
}
\label{fig_comparisions}
\end{figure*}

\begin{table*}
\begin{center}
\caption{Results of quantitively comparing the performance of our method with other state-of-the-art models on the ISPRS benchmark dataset.  Figures in the first nine columns of the table demonstrate the F1 scores respectively for each category, and numbers in the last two columns illustrate the overall accuracy (OA) and average F1 score (Average F1). The boldface text shows the model with the highest performance.}
\label{table_avgF1_comparion}
\resizebox{\textwidth}{!}{
\begin{tabular}{l c c c c c c c c c c c}

\hline
Categories & power & low\_veg & imp\_surf & car & fence\_hedge & roof & facade & shrub & tree & OA & Average F1\\
\hline
UM \citep{horvat2016context} & 0.461 & 0.790 & 0.891 & 0.477 & 0.052 & 0.920 & 0.527 & 0.409 & 0.779 & 0.808 & 0.590 \\
WhuY2 & 0.319 & 0.800 & 0.889 & 0.408 & 0.245 & 0.931 & 0.494 & 0.411 & 0.773 & 0.810 & 0.586 \\ 
WhuY3 \citep{yang2017convolutional} & 0.371 & 0.814 & 0.901 & 0.634 & 0.239 & 0.934 & 0.475 & 0.399 & 0.780 & 0.823 & 0.616 \\
LUH \citep{niemeyer2016hierarchical} & 0.596 & 0.775 & 0.911 & 0.731 & 0.340 & 0.942 & 0.563 & 0.466 & \textbf{0.831} & 0.816 & 0.684 \\
BIJ\_W \citep{wang2018deep} & 0.138 & 0.785 & 0.905 & 0.564 & 0.363 & 0.922 & 0.532 & 0.433 & 0.784 & 0.815 & 0.603 \\
RIT\_1 \citep{yousefhussien2018multi} & 0.375 & 0.779 & 0.915 & 0.734 & 0.180 & 0.940 & 0.493 & 0.459 & 0.825 & 0.816 & 0.633 \\
NANJ2 \citep{zhao2018classifying} & 0.620 & \textbf{0.888} & 0.912 & 0.667 & 0.407 & 0.936 & 0.426 & \textbf{0.559} & 0.826 & \textbf{0.852} & 0.693 \\
WhuY4 \citep{yang2018segmentation} & 0.425 & 0.827 & 0.914 & 0.747 & \textbf{0.537} & \textbf{0.943} & 0.531 & 0.479 & 0.828 & 0.849 & 0.692 \\

D-FCN \citep{wen2020directionally} & 0.704  & 0.802  & 0.914  & \textbf{0.781}  & 0.370  & 0.930  & \textbf{0.605}  & 0.460  & 0.794  & 0.822 & 0.707\\  
\hline
GACNN & \textbf{0.760} & 0.818 & \textbf{0.930} & 0.777 & 0.378 & 0.931 & 0.589 & 0.467 & 0.789 & 0.832 & \textbf{0.715}\\

\hline
\end{tabular}}
\end{center}
\end{table*}

After illustrating the effectiveness of each attention module, we compare our model with other models submitted to the ISPRS 3D Labeling benchmark to exhibit the advantages of the proposed model. The top eight models with the highest performance on the benchmark, including UM \citep{horvat2016context}, WhuY2, WhuY3 \citep{yang2017convolutional}, LUH \citep{niemeyer2016hierarchical}, BIJ\_W \citep{wang2018deep}, RIT\_1 \citep{yousefhussien2018multi}, NANJ2 \citep{zhao2018classifying} and WhuY4 \citep{yang2018segmentation}), are selected for performance comparisons. In addition, we also compare the proposed model with our previous D-FCN model \citep{wen2020directionally}, which has been shown to achieve better classification performance than that of the mentioned eight models. Table \ref{table_avgF1_comparion} displays OA and F1 score of our model and the comparison models on the ISPRS benchmark dataset. Our GACNN model achieves better classification performance in terms of average F1 score than the other models. Specifically, the proposed GACNN model achieves state-of-the-art classification performance for the power line, impervious surfaces categories. 

In addition, Table \ref{table_avgF1_comparion} shows that the previous models, such as the NANJ2 model and D-FCN model,unilaterally achieves good results on a certain evaluation metric (overall accuracy or average F1 score),  but performs poorly on the other indicator. By contrast, our model not only achieves the best performance on average F1 score, but also has a satisfying result on overall accuracy in comparison to other models. Meanwhile, it exhibits difficulty to simultaneously achieve the best overall accuracy and average F1 score on this benchmark dataset, since the number of points in various categories is extremely uneven (see Table \ref{table_Dataset_number}). The categories with a large number of points receive more attention during model training, while the small categories are ignored, which leads to good OA performance but poor average F1 scores. On the other hand, performance enhancement on average F1 score will require the model to learn representative features on a small amount of data, which easily results in model overfitting and constrains on overall accuracy. Overall, our GACNN model achieves a satisfactory balance between OA and average F1 score.

Similarly, to further intuitively illustrate the classification results of different models, we plot the results of four models in Figure \ref{fig_comparisions}, including RIT\_1, NANJ2, WhuY4, and D-FCN, which acquire the four best classification performance in terms of average F1 score . As shown in the Figure  \ref{fig_comparisions}, the three models (a,b,and c) in the first row misclassify the points of facade category into the shrub and roof categories, which validates our assumption that facade category is difficult to distinguish from other categories (see Section \ref{Classification results}). The main reason these models fail is that they cannot learn effective local neighboring features of the point cloud. Specifically, the RIT\_1 model adopts a 1D fully convolutional network that cannot capture multiscale features, and NANJ2 and WhuY4 model lose 3D spatial information during the transformation of point cloud to 2D feature images. However, D-FCN and GACNN model learn effective local structural features by employing their core modules, thus obtaining better performance than that of the other models. Moreover, although most points in the low vegetation category are misclassified as impervious surfaces by all the models, our GACNN model still achieves an acceptable result. 

Moreover, to fairly justify the superiority of our model, we compare our model with other prevalent point cloud deep learning models in the field of computer vision, including PointNet++ \citep{qi2017pointnet++}, PointSIFT \citep{jiang2018pointsift}, PointCNN \citep{li2018pointcnn}, PointCNN with A-XCRF model \citep{arief2019addressing}, KPConv\citep{thomas2019kpconv}, DGCNN\citep{wang2019dynamic} and GACNet\citep{wang2019graph} (see Table \ref{table_avgF1_comparion2}). It can be seen from Table \ref{table_avgF1_comparion2} that the proposed GACNN model have achieved better performance than these advanced point cloud deep learning models, which again demonstrate the superiority of our model. Please note that PointCNN with A-XCRF model is a post-processing method to address the overfitting of PointCNN, and our model has not adopted any post-processing technology. Furthermore, compared to DGCNN and GACNet, which are closely related to our work and recently proposed in the field of computer vision, our model achieves better accuracy and F1 score by a large margin. It demonstrates that adding attention mechanisms does improve model performance and the two attention mechanisms employed in this paper are more effective, obtaining superior classification performance than GACNet with only a local spatial attention mechanism.

\begin{table*}[t]
\begin{center}
\caption{The results of quantititively comparing our method with other prevalent point cloud deep learning models on the ISPRS benchmark dataset. The numbers in the first nine columns of the table show the F1 scores for each category, and the last two columns show the overall accuracy (OA) and average F1 score (Average F1). The boldface text indicates the model with the best performance.}
\label{table_avgF1_comparion2}
\resizebox{\textwidth}{!}{
\begin{tabular}{l c c c c c c c c c c c}
\hline
Categories & power & low\_veg & imp\_surf & car & fence\_hedge & roof & facade & shrub & tree & OA & Average F1\\
\hline

PointNet++ \citep{qi2017pointnet++} & 0.579  & 0.796  & 0.906  & 0.661  & 0.315  & 0.916  & 0.543  & 0.416  & 0.770  & 0.812  & 0.656 \\
PointSIFT \citep{jiang2018pointsift} & 0.557 & 0.807 & 0.909 & \textbf{0.778} & 0.305 & 0.925 & 0.569 & 0.444 & 0.796 & 0.822 & 0.677 \\
PointCNN \citep{li2018pointcnn} & 0.615  & \textbf{0.827}  & 0.918  & 0.758  & 0.359  & 0.927  & 0.578  & 0.491  & 0.781  & 0.833  & 0.695 \\
PointCNN+A-XCRF \citep{arief2019addressing} & 0.630  & 0.826  & 0.919  & 0.749  & \textbf{0.399}  & \textbf{0.945}  & 0.593  & \textbf{0.508}  & \textbf{0.827}  & \textbf{0.850}  & 0.711 \\
KPConv \citep{thomas2019kpconv} & 0.631  & 0.823  & 0.914  & 0.725  & 0.252  & 0.944  & \textbf{0.603}  & 0.449  & 0.812  & 0.837  & 0.684 \\
DGCNN \citep{wang2019dynamic} & 0.676 & 0.804 & 0.906 & 0.545 & 0.268 & 0.898 & 0.488 & 0.415 & 0.773 & 0.810 & 0.641 \\
GACNet \citep{wang2019graph} & 0.628 & 0.819 & 0.908 & 0.698 & 0.252 & 0.914 & 0.562 & 0.395 & 0.763 & 0.817 & 0.660 \\
\hline
GACNN & \textbf{0.760} & 0.818 & \textbf{0.930} & 0.777 & 0.378 & 0.931 & 0.589 & 0.467 & 0.789  & 0.832  & \textbf{0.715} \\
\hline

\end{tabular}}
\end{center}
\end{table*}

{\color{black}{
Furthermore, to analyze the computational cost of our model, we compare the per sample inference time of the proposed model with recent state-of-the-art models. As shown in Table \ref{table_computational_cost}, it can be noticed that the inference time of our model is close to the comparing models. Our model achieves new state-of-the-art classification performance with a promising computation efficiency.
}
\begin{table}[h]
\begin{center}
\caption{Model inference time comparison between our model and recent state-of-the-art models on the ISPRS benchmark dataset. }
\label{table_computational_cost}
\begin{tabular}{l c}
\hline
Model & Inference Time
\\
\hline
PointNet++\citep{qi2017pointnet++} & \textbf{0.140s} \\
PointSIFT\citep{jiang2018pointsift} & 0.389s \\
PointCNN\citep{li2018pointcnn} & 0.220s \\
DGCNN\citep{wang2019dynamic} & 0.379s \\
GACNet\citep{wang2019graph} & 0.952s \\
\hline
GACNN & 0.379s \\
\hline
\end{tabular}
\end{center}
\end{table}
}

\subsection{The generalization capability of the proposed model}

\begin{figure*}[h]
\centering
\includegraphics[width=17cm, height=10cm]{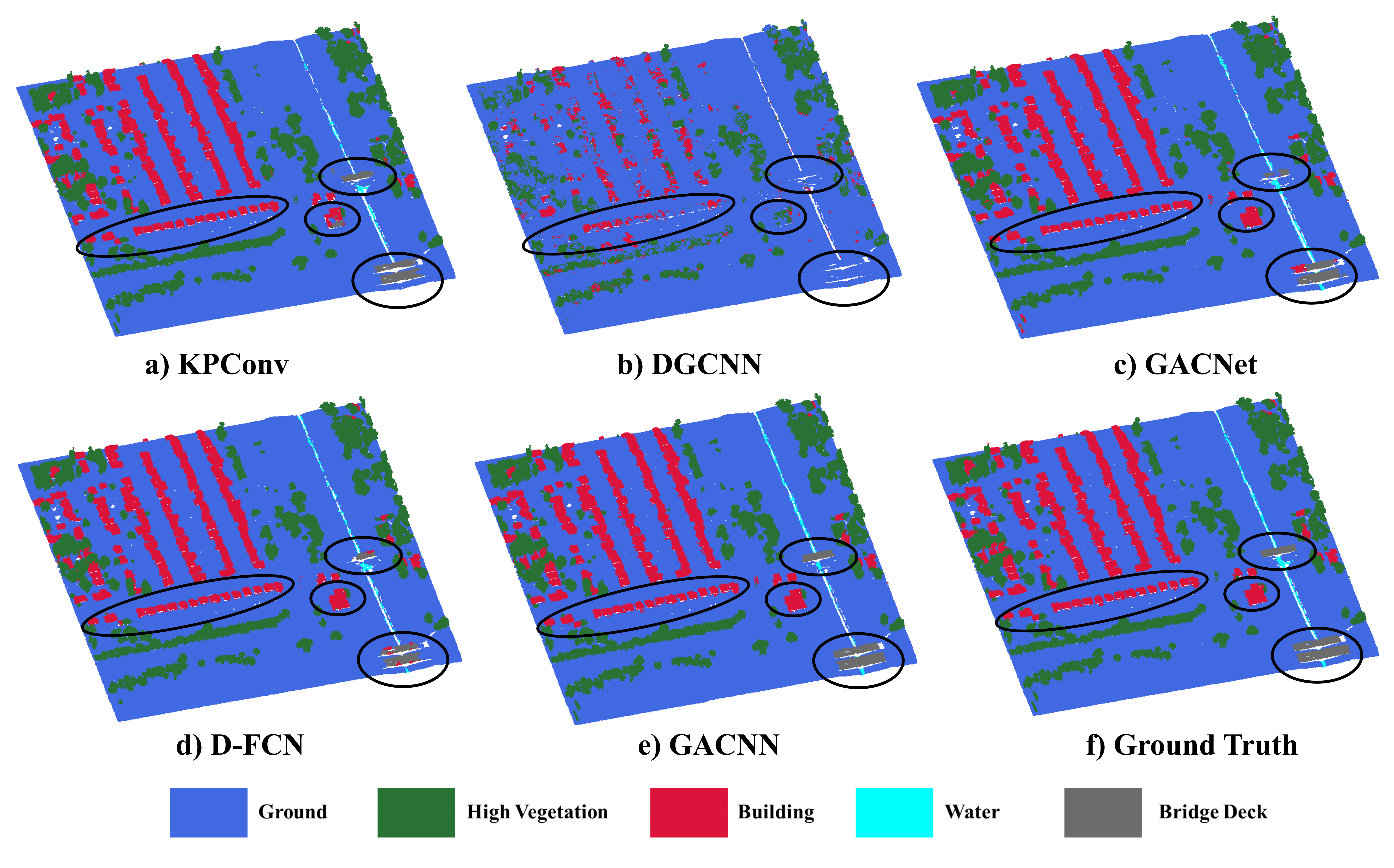}
\caption{The classification results of KPConv model, DGCNN model, GACNet model, D-FCN model and our proposed GACNN model applied to a selected complicated test scene of 2019 Data Fusion Contest Dataset. The black circle indicates the most obvious differences of classification results obtained by the six models. Best viewed in color.}
\label{fig_dfc_Testset_compare}
\end{figure*}

\begin{table}[h]
\begin{center}
\caption{The number of points and detailed category distribution of Data Fusion Contest Dataset}
\label{table_dfc_Dataset_number}
\begin{tabular}{l c c c}
\hline
Categories & Training set & Test set & Total
\\
\hline   
Ground & 48,069,788 & 6,148,920 & 54,218,708\\
High Vegetation & 10,775,292 & 1,321,488 & 12,096,780\\
Building & 10,269,771 & 782,455 & 11,052,226\\
Water & 1,356,899 & 19,266 & 1,376,165\\
Bridge Deck & 934,545 & 25,034 & 959,579\\
\hline
Total & 71,406,295 & 8,297,163 & 79,703,458\\       
\hline
\end{tabular}
\end{center}
\end{table}

\begin{table}[h]
\begin{center}
\caption{The classification overall accuracy (OA) and average F1 score of our proposed GACNN model with different attention modules on the 2019 Data Fusion Contest Dataset. The boldface text indicates the model with the highest performance.}
\label{table_effect_attention_2}
\begin{tabular}{l c c}
\hline
Model & OA & Average F1
\\
\hline
a) GACNN(w/o global, w/o local) & 0.937 & 0.798 \\
b) GACNN(w/ global, w/o local)  & 0.945 & 0.820 \\
c) GACNN(w/o global, w/ local) & 0.941 & 0.817 \\
d) GACNN(w/ global, w/ local\_edge) & 0.947 & 0.826 \\
e) GACNN(w/ global, w/ local\_density) & 0.948 & 0.823 \\
\textbf{f) GACNN(w/ global, w/ local)} & \textbf{0.951} & \textbf{0.828} \\
\hline
\end{tabular}
\end{center}
\end{table}

\begin{table*}[htb]
\begin{center}
\caption{The results of quantitively comparing our method with other prevalent models on the 2019 Data Fusion Contest Dataset. Figures in the first nine columns of the table show the F1 scores separately for each category, and numbers in the last two columns illustrate the overall accuracy (OA) and average F1 score (Average F1). The boldface text indicates the model with the highest performance.}
\label{table_dfc_avgF1_comparion}
\begin{tabular}{l c c c c c c c c c c c}

\hline
Categories & Ground & High Vegetation & Building & Water & Bridge Deck & OA & Average F1\\
\hline
PointNet++ \citep{qi2017pointnet++} & 0.983  & 0.958  & 0.797  & 0.044  & 0.073  & 0.927  & 0.571 \\
PointSIFT \citep{jiang2018pointsift} & 0.986  & 0.970  & 0.855  & \textbf{0.464}  & 0.604  & 0.940  & 0.776  \\
PointCNN \citep{li2018pointcnn} & 0.987  & 0.972  & 0.849  & 0.441  & 0.653 & 0.938 & 0.780\\
KPConv \citep{thomas2019kpconv} & 0.984  & 0.942  & 0.874  & 0.430  & \textbf{0.775}  & 0.945  & 0.801  \\
DGCNN \citep{wang2019dynamic} & 0.982 & 0.953 & 0.745 & 0.112 & 0.284 & 0.929 & 0.615 \\
GACNet \citep{wang2019graph} & 0.985 & 0.968 & 0.852 & 0.403 & 0.687 & 0.937 & 0.779 \\
D-FCN \citep{wen2020directionally} & 0.991  & \textbf{0.981}  & 0.899  & 0.450  & 0.730  & \textbf{0.956}  & 0.810 \\
\hline
GACNN & \textbf{0.993}  & 0.968  & \textbf{0.911}  & 0.425  & \textbf{0.844}  & 0.951  & \textbf{0.828}\\  

\hline

\end{tabular}
\end{center}
\end{table*}

To demonstrate the generalization capability of the proposed model, we further conduct experiments on 2019 Data Fusion Contest Dataset. The 2019 Data Fusion Contest, organized by the Image Analysis and Data Fusion Technical Committee (IADF TC) of the IEEE Geoscience and Remote Sensing Society (GRSS), the Johns Hopkins University (JHU), and the Intelligence Advanced Research Projects Activity (IARPA), provides a large-scale Urban Semantic 3D (US3D) dataset covering approximately 100 square kilometers over Jacksonville, Florida and Omaha, Nebraska, United States. The point clouds in US3D dataset are labeled as five categories, including ground, trees, buildings, water and bridge, and are stored in ASCII text files with format {x, y, z, intensity, return number}.

This training set of Contest Dataset consists of 110 regions, from which the 100 regions are selected for the training dataset and the other 10 regions are used for testing dataset in this experiment. The number of points in this dataset and detailed category distribution is shown in Table \ref{table_dfc_Dataset_number}. Similarly, we randomly select a 128m*128m*210m cuboid region from each region during training and arbitrarily choose 8192 points from the cuboid as model input. As for the testing stage, each region of the testing dataset is split into blocks of 128m*128m grids in the horizontal direction and all points is segmented blocks are input into the model for prediction. Note that the architecture and parameters of our model remain unchanged.

{\color{black}
We first evaluate the effectiveness of the proposed graph attention convolution module and conduct an ablation experiment similar to what we have done in Section \ref{ab_attention}. The classification results of the six models which comprise of different attention modules are listed in Table \ref{table_effect_attention_2}, from which it can be found that the proposed local and global attention module do help to improve the classification performance to some extent as we discussed in Section \ref{ab_attention}.
}

The PointNet++ \citep{qi2017pointnet++}, PointSIFT \citep{jiang2018pointsift}, PointCNN \citep{li2018pointcnn}, KPConv \citep{thomas2019kpconv}, DGCNN \citep{wang2019dynamic}, GACNet \citep{wang2019graph} and DFCN model \citep{wen2020directionally} are selected as comparison model, and the classification results on test regions are shown in Table \ref{table_dfc_avgF1_comparion}. From Table \ref{table_dfc_avgF1_comparion}, it can be seen that our model obtains the best classification performance on average F1 score and achieves the highest F1 score for three of the five categories, including ground, building and bridge deck. Moreover, Figure \ref{fig_dfc_Testset_compare} displays the classification results of the four latest models among above prevalent point cloud deep learning models on a randomly selected test region, from which we can find that KPConv model and our model classify the points of bridge deck category more correctly. But there are some points of building category are misclassified by KPConv model. At the same time, from Table \ref{table_dfc_avgF1_comparion} and Figure \ref{fig_dfc_Testset_compare}, it can be found that our model, which incorporates local structural features and global contextual information simultaneously, performs better than DGCNN model and GACNet model, two closely related works in the field of computer vision, which further indicates the effectiveness of the proposed attention modules. As a result, the proposed model still outperforms other state-of-the-art models on the 2019 Data Fusion Contest Dataset and demonstrates a favorable generalization capability.

\section{Conclusions}\label{Conclusion}

In this paper, a graph attention convolution neural network (GACNN) that is directly applied to unstructured 3D point clouds is proposed to conduct airborne LiDAR point cloud classification. The GACNN model is an end-to-end encoder-decoder network developed based on the proposed graph attention convolution module, which consists of a global attention module and a local attention module. These two attention modules take local structural features and global contextual information into consideration respectively, and their effectiveness is validated through a set of comparative experiments. Specifically, the proposed graph attention convolution module is capable of dynamically learning convolution weights according to the local structure of the point cloud, considering the unbalanced density distribution of the point cloud, and at the same time paying attention to spatial relationships among all points in global. Moreover, we compare our GACNN model with other state-the-of-art models on both ISPRS 3D Labeling Dataset and 2019 Data Fusion Contest Dataset, the results of which demonstrate the proposed model is superior to most of prevalent point cloud classification models, whether in the field of photogrammetry and remote sensing or computer vision, and achieves a new state-of-the-art classification performance in terms of average F1 score. Moreover, experiments on two different datasets show the favorable generalization capability of our model.

\section*{ACKNOWLEDGEMENTS (Optional)}\label{ACKNOWLEDGEMENTS}
This work was supported by the Beijing Municipal Science and technology Project (grant No. Z191100001419002). Additionally, we would like to gratefully acknowledge the ISPRS for providing airborne LiDAR data.

{
\begin{spacing}{0.9}
\bibliography{bibliography} 
\end{spacing}
}



\end{document}